\documentclass[conference]{IEEEtran}
\usepackage{cite}
\usepackage{amsmath,amssymb,amsfonts}
\usepackage{algorithmic}
\usepackage{graphicx}
\usepackage{textcomp}

\usepackage{amsbsy}
\usepackage{booktabs}
\usepackage[english]{babel}
\usepackage[utf8]{inputenc}
\usepackage{csquotes}
\usepackage{interval}
\usepackage{microtype}
\usepackage{multirow}
\usepackage{siunitx}
\usepackage{tabularx}
\usepackage{url}

\DeclareMathOperator*{\argmin}{\arg\!\min}
\def\realnumbers{\mathbb{R}}

\intervalconfig{soft open fences}

\begin{document}

\title{Efficient Optimization of Echo State Networks for Time Series Datasets}

\author{\IEEEauthorblockN{Jacob Reinier Maat\IEEEauthorrefmark{1},
Nikos Gianniotis\IEEEauthorrefmark{2} and
Pavlos Protopapas\IEEEauthorrefmark{1} 
\IEEEauthorblockA{\IEEEauthorrefmark{1}Institute For Applied Computational Science\\
Harvard University\\
Cambridge, MA 02138\\
Email: maat@alumni.harvard.edu\\
Email: pavlos@seas.harvard.edu}}
\IEEEauthorblockA{\IEEEauthorrefmark{2}Heidelberg Institute for Theoretical Studies\\
69118 Heidelberg, Germany\\
Email: nikos.gianniotis@h-its.org}}

\maketitle

\begin{abstract}
Echo State Networks (ESNs) are recurrent neural networks that only train their output layer, thereby precluding the need to backpropagate gradients through time, which leads to significant computational gains. Nevertheless, a common issue in ESNs is determining its hyperparameters, which are crucial in instantiating a well performing reservoir, but are often set manually or using heuristics. In this work we optimize the ESN hyperparameters using Bayesian optimization which, given a limited budget of function evaluations, outperforms a grid search strategy. In the context of large volumes of time series data, such as light curves in the field of astronomy, we can further reduce the optimization cost of ESNs. In particular, we wish to avoid tuning hyperparameters per individual time series as this is costly; instead, we want to find ESNs with hyperparameters that perform well not just on individual time series but rather on groups of similar time series without sacrificing predictive performance significantly. This naturally leads to a notion of clusters, where each cluster is represented by an ESN tuned to model a group of time series of similar temporal behavior. We demonstrate this approach both on synthetic datasets and real world light curves from the MACHO survey. We show that our approach results in a significant reduction in the number of ESN models required to model a whole dataset, while retaining predictive performance for the series in each cluster.
\end{abstract}

\section{Introduction} \label{intro}
Echo State Networks (ESNs) \cite{Jaeger2001} are an instance of recurrent neural networks (RNNs) whose distinctive feature is that they only adapt their output layer in training, while keeping the weights of recurrent and input connections fixed. 
This avoids propagating gradients back through time, which is computationally expensive and fraught with numerical difficulties \cite{pascanu2013}.
The ESN architecture is governed by numerous hyperparameters that greatly influence its predictive performance \cite{Lukosevicius2012a}. 
These parameters govern the properties of its reservoir, i.e. the hidden nodes that make up the recurrent part of the network. 
Depending on the specification of the ESN, the number of parameters can range from 4 \cite{Rodan2010} to 7 \cite{Jaeger2001}. 
ESN performance depends critically on this parametrization. 

Current approaches to optimization of these hyperparameters are typically slow or suboptimal, and include setting them manually based on experience \cite{Lukosevicius2012a}, heuristic search \cite{Jiang2008}, gradient-based optimization \cite{Jaeger2007} and grid search \cite{Rodan2010}. 
While rules of thumb and heuristics are useful in practice, these rules do not guarantee optimality of the hyperparameters for an individual task. 
Gradient-based optimization is computationally expensive and only works for the subset of the parameters that have defined gradients. 
This is non-trivial, because parameters governing the size and connectivity of a reservoir have a direct effect on the memory capacity of the model and, therefore, its performance \cite{Farkas2016}. Additionally, since prior research has shown that the optimization problem is not guaranteed to be convex \cite{Jaeger2007}, gradient descent can lead to suboptimal, local optima.

The most exhaustive approach to hyperparameter optimization so far is a grid search strategy, which finds the optimal parameter setting by evaluating all combinations on a discretized grid. While this is suboptimal due to discretization of the space, it may come close to the optimal setting if the resolution of the grid is high enough, albeit this can make the optimization extremely expensive. In addition, the curse of dimensionality makes a grid search infeasible for high dimensional spaces.

In this work we reduce the burden of ESN optimization twofold, so to increase the model's practicality in large-scale applications. 
Firstly, we use Bayesian optimization to optimize the hyperparameters governing the reservoir and show this outperforms a grid search strategy in terms of computational cost, while retaining the ability to optimize all parameters, including discrete ones. 

Secondly, in the context of large scale time series datasets it may not even be required to model every time series individually -- each with its own hyperparameters and readout weights. Often the data generating processes are limited to a number of (unknown) latent classes. For example, in astronomical sky surveys, light curves (photometry measured over time) greatly outnumber the number of stellar classes. We use this premise of redundancy in large scale time series dataset to optimize groups of similar time series using cluster-based ESNs optimized by Bayesian optimization, rather than individual series. This leads to another speedup by greatly reducing the number of models needing to be trained an optimized.

\subsection*{Related Work}
While \cite{yperman2016} have (concurrently) shown it is possible to use Bayesian optimization to optimize hyperparameters in ESNs, they do not provide evidence that this is better than the commonly used grid-search strategy. In this work we compare these strategies in terms of required function evaluations, and show that Bayesian optimization is more efficient than a grid search on benchmark sets. In addition, we show that the technique can be efficiently used to cluster time series by their underlying dynamics, which is useful in domains like astrophysics where large sets of unlabeled time series need to be grouped in some fashion.


Section \ref{background} will go into the background theory behind Echo State Networks and Bayesian optimization.
In section \ref{bo_esn} we show how Bayesian optimization can be practically applied to the hyperparameter optimization problem of an ESN. 
Section \ref{clustering} uses this approach show how Bayesian optimized ESNs can be used as the to effectively group time series into clusters.
In section \ref{experiments} we show that Bayesian optimization significantly outperforms a grid search in terms of model evaluations, and we show how ESNs optimized using Bayesian Optimization can be used as cluster centers in clustering families of time series.
This is demonstrated in numerical experiments on synthetic as well as real-world data.

\section{Background} \label{background}

\subsection{Echo State Networks}

\begin{figure}
	\centering
	\includegraphics[width=0.9\linewidth]{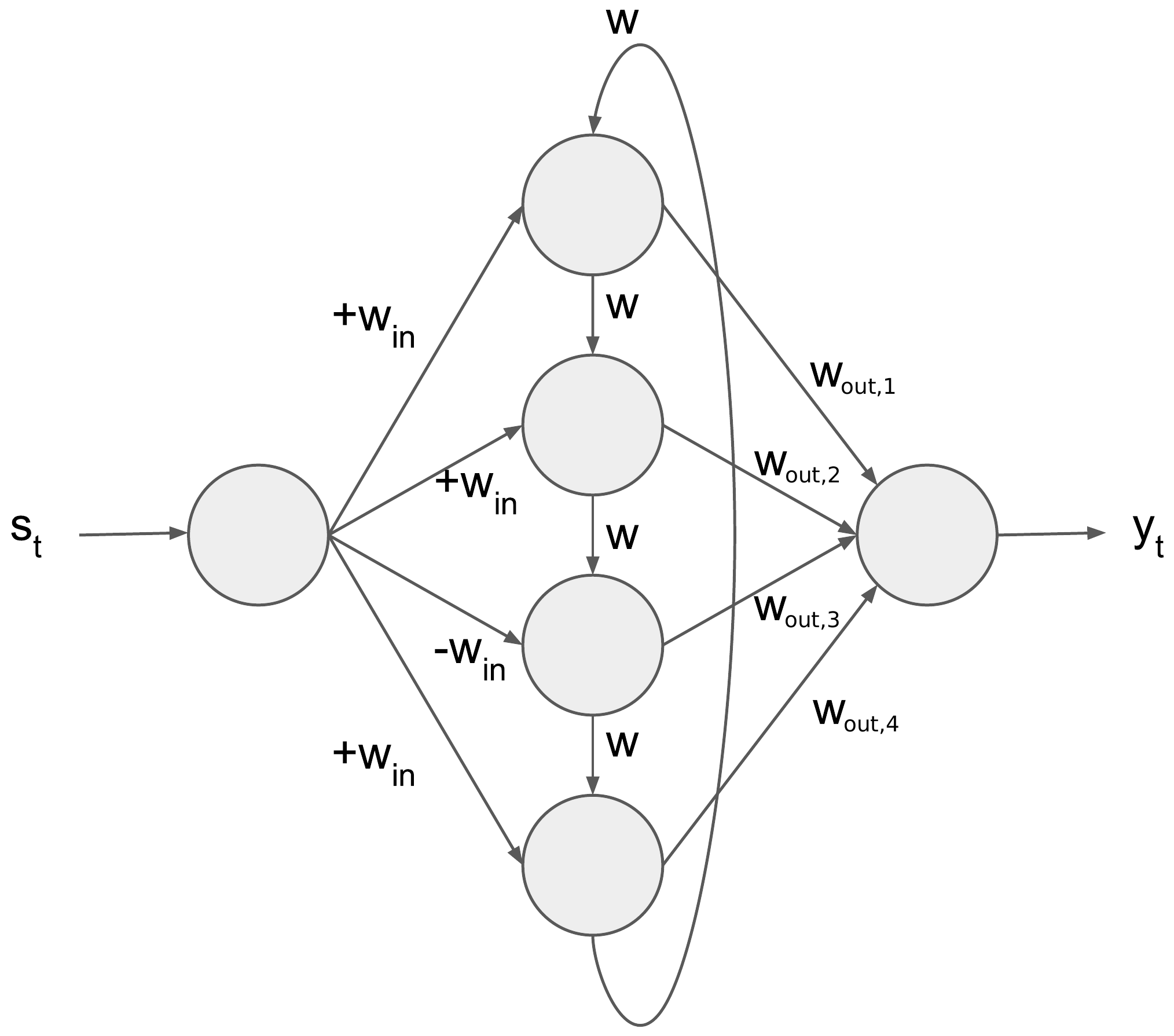}
	\caption{Sketch of Simple Cyclic Reservoir (SCR) for one-dimensional time series. Input weights are parameterized by scalar $|w_{in}|$ whose signs are deterministically generated. Recurrent weights are parameterized by the same value $w$. Only the readout weights $\mathbf{W_\mathrm{out}}$ are adapted during training.}
	\label{figure:scr_esn}
\end{figure}
ESNs are discrete time, deterministic state-space models in which only the output, i.e. readout, layer is trained, while the  input and hidden layer are generated stochastically \cite{Jaeger2001}. ESNs predict a target $\mathbf{y}(t) \in \realnumbers^{Y}$ given input series $\mathbf{s}(t) \in \realnumbers^{S}$, where $t \in [1, T]$, $T$ is the length of the time series, $S$ is the number of input series, and $Y$ is the number of output series being predicted. The hidden recurrent part of the ESN, referred to as the reservoir, contains $N$ neurons and is viewed as a fading memory whose current state $\mathbf{x}(t)$ summarizes previously processed inputs. At time $t$, the ESN sees an input $\mathbf{s}(t)$ and updates its state via\footnote{We pose no feedback connections from the output to the reservoir.}:
\begin{equation}
\small
	\mathbf{x}(t) = \tanh{(\mathbf{W}_\mathrm{in}[1;\mathbf{s}(t)] + \mathbf{W}\mathbf{x}(t-1))} \ ,
\label{eq:state_update}
\end{equation}
\noindent where $\mathbf{W}_\mathrm{in}\in \realnumbers^{N \times S}$ 
are the inputs weights and $\mathbf{W} \in \realnumbers^{N \times N}$ are the weights of the hidden recurrent layer, i.e. the reservoir.
The ESN makes a prediction for the target at time $t$ via:
\begin{equation}
\small
	\hat{\mathbf{y}}(t) = [\mathbf{1}; \mathbf{x}(t)]\mathbf{W_\mathrm{out}} \ ,
\end{equation}
where the concatenated column of ones acts as bias, and $\mathbf{W_\mathrm{out}}\in \realnumbers^{(N+1)\times Y}$ are the output, or readout, weights of the network.

ESNs are constructed in two stages.
Firstly, the inputs weights $\mathbf{W}_\mathrm{in}$
and hidden weights $\mathbf{W}$ are stochastically generated  and fixed \cite{Jaeger2001}.
In a second stage, the readout weights $\mathbf{W_\mathrm{out}}$ are optimized. Starting with an initial arbitrary state, e.g. ${\mathbf{x}}(0)={\mathbf{0}}$, for each input $\mathbf{s}(t)$ the state activations ${\mathbf{x}}(t)$, calculated via \eqref{eq:state_update}, are recorded row-wise in matrix\footnote{Typically, some initial states are discarded 
in order to eliminate dependence on the initial arbitrary state \cite{Jaeger2001}.} $\mathbf{X}\in\realnumbers^{T\times N}$.

Secondly, having collected all states in $\mathbf{X}$ and given
the corresponding targets as $\mathbf{y}$, $\mathbf{W_\mathrm{out}}$
is obtained by minimizing the least squares problem:
\begin{equation}
\small
\sum_{t=1}^T (\mathbf{y}(t) - \hat{\mathbf{y}}(t) )^2   + \lambda\|\mathbf{W_\mathrm{out}} \|^2  \ ,
\label{eq:loss_esn}
\end{equation}
where  $\lambda\geq 0$ is a regularization parameter.

\subsection{Simple Cyclic Reservoir}

A practical difficulty when working with the ESN is the fact that its performance depends
on the random construction of the network topology and its hidden (reservoir) weights. The Simple Cyclic Reservoir (SCR) was put forward as a way of constructing the ESN in a fully deterministic way \cite{Rodan2010} without compromising performance. The SCR connects the hidden neurons in a cycle and simplifies the weight structure: all elements in $\mathbf{W}_\mathrm{in}$ are set to the same absolute value of a scalar $w_\mathrm{in} \in [0.0,1.0]$. The signs of the elements in $\mathbf{W}_\mathrm{in}$ are deterministically set by a pseudo-random sequence. All elements in the reservoir matrix $\mathbf{W}$ are set to $0$ apart from the elements in the lower sub-diagonal and the upper right corner which are also set to the same scalar, i.e. $\mathbf{W}_{n+1,n} = w, n\in[1,\dots,N-1]$ and $\mathbf{W}_{1,N}=w$ where $w \in [0.0,1.0]$. A sketch of the SCR architecture is displayed in Fig. \ref{figure:scr_esn}.

Training proceeds in exactly the same way as for the standard ESN. Table \ref{table:hyperparameters} lists the hyperparameters associated an SCR.
We collectively denote all hyperparameters by $\mathbf{\theta}$, and make them explicit in the predictions of the SCR with the notation $\hat{\mathbf{y}}_{\mathbf{\theta}}(t)$.

\begin{table}
    \centering
    \caption{SCR hyperparameters optimized using Bayesian Optimization}
	\begin{tabular}{ll}
      \toprule
      Parameter 		& Description \\
      \midrule
      $N$ 				& Number of hidden neurons in reservoir \\
      $w_\mathrm{in}$   & Scaling of inputs \\
      $w$   			& Weight connecting the nodes in SCR \\
      $\lambda$			& Regularization parameter (see eq. \eqref{eq:loss_esn}) \\
      \bottomrule
	\end{tabular}
    \label{table:hyperparameters}
\end{table}

\subsection{Bayesian Optimization}

Bayesian optimization (BO) is a gradient-free global optimization method to optimize arbitrary functions \cite{Snoek2012}, and is one way to perform surrogate-based optimization \cite{queipo2005}. It was introduced to minimize loss functions $f(\mathbf{\theta})$ of machine learning models whose (hyper)parameters $\mathbf{\theta}$ are difficult to tune, such as when the optimization problem exhibits many local optima, or when gradients are unavailable.

BO, as formulated in \cite{Snoek2012}, treats the loss $f(\mathbf{\theta})$ as a latent function to be inferred, and models it with a Gaussian Process (GP), i.e. it maintains a probability distribution $\mathrm{p_{cur}}(f)$ over an infinite set of candidate functions of what the true loss function $f(\mathbf{\theta})$ might be. The true loss is only known at the parameters $\mathbf{\theta}$ where it has been evaluated. Given the current $\mathrm{p_{cur}}(f)$, BO samples a new $\mathbf{\theta}$ which it considers as a likely candidate for the sought minimum (acquisition). The newly sampled $\mathbf{\theta}$ is viewed as a new observation from $f(\mathbf{\theta})$ and is used to update the current distribution over functions, i.e. we get an updated posterior $\mathrm{p_{new}}(f) \propto \mathrm{p}(\mathbf{\theta}|f) \mathrm{p_{cur}}(f)$ compatible with all $\mathbf{\theta}$'s evaluated so far.

An important element in BO is the acquisition function that decides which  $\mathbf{\theta}$ to query next. Amongst the various acquisition functions available, we use the Lower Confidence Bound (LCB) \cite{Snoek2012},
which chooses to query the loss at $\argmin_{\mathbf{\theta}} \mathop{E}[f(\mathbf{\theta})] - \kappa \cdot \mathop{\sigma}[f(\mathbf{\theta})]$. The scalar  $\kappa \geq 0$ balances exploitation (querying where the expectation of a minimum is high) and exploration (querying where the standard deviation of the loss is high).

Typically, BO starts off by evaluating the loss $f(\mathbf{\theta})$ on a randomly picked set of $\mathbf{\theta}$'s for an initial exploration of what the landscape of $f(\mathbf{\theta})$ looks like. This is useful for initializing the  hyperparameters of the GP (e.g. kernel length scales). Thereafter, queries on $\mathbf{\theta}$ are decided via the acquisition function, while GP hyperparameters are updated after each query. Convergence is reached when the next query for $\mathbf{\theta}$ falls within a radius $\epsilon$ of the previous query.

\section{Bayesian Optimization applied to Echo State Networks} \label{bo_esn}

In our case, we use BO to fit a GP (the surrogate) to the prediction error of an SCR (our model) as a function of its hyperparameters. We then iteratively optimize its hyperparameters by optimizing an acquisition function, as defined above.

We optimize the parameters $\mathbf{\theta}=[ N, w_\mathrm{in}, w, \lambda]$ of an SCR in the task of predicting the next time step of a time series $\mathbf{y}$. Whereas gradient-based optimization methods cannot optimize parameters that are not readily expressible in gradients, such as $N$, BO can optimize all parameters using the LCB criterion to guide the search.

As mentioned before, typically the objective in \eqref{eq:loss_esn} (or variations thereof) is optimized for obtaining the readout weights $\mathbf{W_\mathrm{out}}$ of an SCR on a time series $\mathbf{y}$.
In principle, the same objective could be used by BO to optimize the hyperparameters $\mathbf{\theta}$. The joint optimization of $\mathbf{W_\mathrm{out}}$ and $\mathbf{\theta}$ could be done by
iteratively alternating between two steps: in the first one, $\mathbf{\theta}$ is kept fixed and $\mathbf{W_\mathrm{out}}$ is optimized, and in the second step we keep fixed $\mathbf{W_\mathrm{out}}$ while optimizing $\mathbf{\theta}$ with BO.
However, optimizing both hyperparameters and parameters can lead to overfitting. Therefore, we suggest a more robust training objective based on a $K$-fold cross-validated likelihood \cite{smyth2000model}.

Specifically: we split the set of time steps $\mathbf{t}~=~\{1,2,\dots,T\}$ indexing time series $\mathbf{y}$ into $K$ equal-sized sets denoted by $\mathbf{t}_k$. 
In each of the $K$ rounds, the SCR is trained on the $K-1$ sets of indices 
by minimizing $\sum_{t\in \mathbf{t}\setminus \mathbf{t}_k} (\mathbf{y}(t) -  \hat{\mathbf{y}}_{\theta}(t) )^2+ \lambda\|\mathbf{W_\mathrm{out}} \|^2$,
and  
we measure its performance on how well it predicts on the left-out time steps $\mathbf{t}_k$ as  $\sum_{t\in \mathbf{t}_k} (\mathbf{y}(t) -  \hat{\mathbf{y}}_{\theta}(t) )^2$.
Thus, we define the objective for the BO as the average over the left-out performances:
\begin{equation}
\small
f(\theta) = \frac{1}{K} \sum_{k=1}^K \sum_{t\in \mathbf{t}_k} (\mathbf{y}(t) -  \hat{\mathbf{y}}_{\theta}(t) )^2 \ .
\label{eq:objective_for_bo}
\end{equation}
In the optimization, we initialize the GP that models $f(\mathbf{\theta})$ with a zero mean function (since we center the samples from $f(\theta)$ every iteration), an ARD Matérn 5/2 kernel covariance matrix, and an LCB acquisition function with $\kappa = 2$, a configuration that has empirically been proven a good default choice across a variety of problems \cite{Snoek2012}. The optimization is performed in log space.

\section{Clustering Families of Time Series} \label{clustering}

As mentioned before, a main concern of our work is reducing the computational cost associated with optimizing the ESN hyperparameters $\mathbf{\theta}$. 
Typically given a dataset of $n$ time series $\{\mathbf{y}_1,\dots,\mathbf{y}_n \}$, one determines individual hyperparameters $\theta$ and outputs weights $\mathbf{W}_{\mathrm{out}}$ for each time series $\mathbf{y}_i$ in the dataset. In many large scale time series databases settings this is not strictly needed, because it is known that the series are sourced from only a small number of generating processes. For example, an astronomical sky survey will deliver time series (light curves) that all belong to one particular class of stars with their own dynamics (albeit it is not initially known to which class they belong). We take advantage of this redundancy and avoid unnecessary (expensive) computation, by finding ESNs that work well not just for an individual time series but rather for multiple time series of similar temporal behavior, i.e. families of time series. 
To that end, we formulate a clustering approach whereby each cluster is characterized by an SCR that exhibits good predictive performance for a family of time series.

Time-series are softly assigned to clusters, and each cluster's SCR is then adapted to fit its assigned members, akin to fuzzy clustering approaches \cite{nayak2015fuzzy}. Here adaptation involves adapting the readout weight vector of the SCR jointly with the hyperparameters governing the reservoir. We use BO to optimize the cluster dependent SCRs.

Let $C$ be the number of clusters and $n$ the number of time series $\mathbf{y}_{i}, i\in[1,n]$ to be clustered. Each cluster is characterized by a
SCR parametrized by a readout $\mathbf{W}_\mathrm{out,c}$ and hyperparameters $\mathbf{\theta}_{c}$. Inspired by fuzzy clustering, we calculate memberships via the softmax function as:
\begin{equation}
\small
m_{ic} = \frac{e^{- f_i(\mathbf{\theta}_c) }}{\sum_{c'=1}^C e^{ -f_i(\mathbf{\theta}_{c'})}} \ ,
\end{equation}
where $f_i(\mathbf{\theta}_c)$ is the objective in \eqref{eq:objective_for_bo} expressing the performance of the $c$-th SCR on the $i-$th time series.

Given the memberships, the total loss function for the fuzzy-clustering model reads:
\begin{equation}
\small
\sum_{i=1}^n \sum_{c=1}^C m_{ic} \ f_i(\mathbf{\theta}_c) \ , 
\label{eq:loss_for_fuzzy_clustering}
\end{equation}

Keeping the memberships $m_{ic}$ fixed, we then optimize every cluster model via BO in \eqref{eq:loss_for_fuzzy_clustering} independently of all others (their parameters $\mathbf{\theta}_c$ do not interact). The Bayesian optimization of every SCR and the membership calculation step are repeated alternately until the total validation error stops decreasing (or starts increasing) for each series' most probable model. This criterion is formalized as:
\begin{equation}
\small
e_l = \sum_{i=1}^n \min_c f_i(\mathbf{\theta}_c) \ .
\end{equation}
This is an early stopping strategy, meaning we stop clustering when $e_l > e_{l-1}$, where $l$ denotes the $l$-th iteration in clustering. Even though early stopping does not guarantee global optimality, it speeds up the clustering process and helps to avoid overfitting by stopping when generalization error is lowest \cite{prechelt1998}.

\section{Numerical experiments} \label{experiments}

We describe the synthetic and real-world datasets used in the numerical experiments\footnote{The SCR was implemented in Python and is available on GitHub. For the implementation of the Bayesian optimization method we adapted the package {GPyOpt} \cite{gpyopt2016}}. The first set of experiments show how BO can be a good alternative to grid search in optimizing the hyperparameters of an ESN.  
The second set of experiments show how it is possible to reduce the number of ESN models required for modeling  datasets of multiple time series.

\subsection{Synthetic datasets}

\paragraph{Mackey-Glass} This is a chaotic time series that has been extensively used as a benchmark for ESNs e.g. \cite{Jaeger2001}, and is defined as:
\begin{equation}
\small
\frac{dx}{dt} = 0.2 \frac{ x(t - \tau) }{1+{x(t - \tau)}^{10}} - 0.1 x(t) \ .
\end{equation}
Using $\tau=30$ and initial values of $x(0) = 0.5$ and $x(\tau) = 1$, we generate\footnote{All synthetic benchmark series were generated with the {TimeSynth} library for Python \cite{Maat2017}.}
$1500$ samples. To increase the difficulty of the task, white noise is added with a standard deviation of 0.05. Subsequently, we split the series into 1000 samples for the training/validation and 500 samples for the test set. 

\paragraph{NARMA} The second synthetic benchmark is a Nonlinear Autoregressive Moving Average (NARMA) series of the 10th order:
\begin{equation}
\small
\begin{aligned}
\small
	y(k+1) = {} & 0.3 y(k) + 0.05 y(k) \sum_{i=0}^{9} y(k-i) + \\ 
				& 1.5 s(k-9) s(k) + 0.1
\end{aligned}
\end{equation}
where $s(k)$ is an input sequence drawn independently from a uniform distribution, $\mathrm{U}\interval[open]{0}{0.5}$. As initialization $y$ was set to $0$ for $k<0$. Again, we generate $1500$ samples, of which the first 1000 samples are used for training/validation, and the last 500 samples for testing.

\subsection{Real-world datasets}

\paragraph{EMG} The first dataset is an electromyogram (EMG) of a healthy subject \cite{Rutkove2010}, retrieved from PhysioNet \cite{PhysioNet}. EMGs are nerve conduction studies used to assess muscle function, and are used to diagnose disorders like muscular dystrophies and neuropathies \cite{Rutkove2010}. We used 2000 points from this dataset for training and validation, and 500 points for testing.

\begin{figure}
  \centering
  \includegraphics{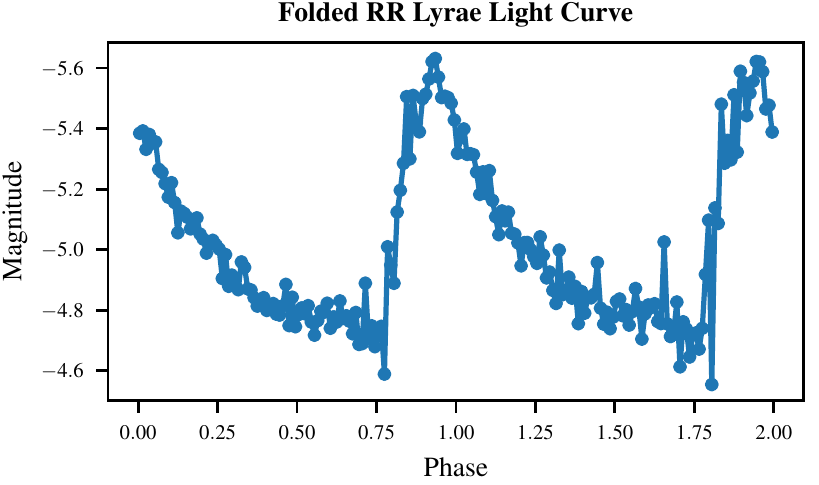}
  \caption{Two unfiltered periods of the RR Lyrae light curve used for the comparison in Table \ref{table:results}}
  \label{fig:RRL}
\end{figure}

\paragraph{MACHO} The second dataset is an astronomical light curve of the RR Lyrae (RRL) class, sourced from the MACHO survey \cite{Macho1996}. RRLs are type of variable star, whose their brightness fluctuates with time, producing a time series. This series uniquely identifies the type of the variable star at hand. An example is shown in Fig. \ref{fig:RRL}. Because of sparse, irregular sampling, light curves are analyzed in phase space by folding them by their period, i.e. we align individual sample points by their position in the light curve's period. We determine the period of our RRL using the MHAoV method \cite{schwarzenberg1996}, as implemented in the package {P4J} \cite{P4J}. The folded series is subsequently binned into regular intervals for usage in ESNs. Data is binned to 500 points covering 10 periods, so to make sure the ESN has enough training data to learn the light curve's structure after discarding initial transient (washout). Any missing values are interpolated quadratically. Outliers are filtered out by applying a Savitzky-Golay filter \cite{Savitzky1964}. Of the 500 points, 400 are used for training and validation, and 100 for testing.

\subsection{Comparison: Bayesian optimization vs. grid search}

\paragraph{Setup} We  demonstrate the efficacy of Bayesian optimization by comparing it against a grid search, and observe the number of function evaluations that it needs to reach the same test error as the grid search. The parameter domains for Bayesian optimization read: number of nodes $N \in \{50,51,\dots,200\}$, cyclic weight $r_c \in \interval[open]{0.01}{0.95}$, input weight $w_\mathrm{in} \in \interval[open]{0.01}{0.95}$ and the L2-regularization parameter  $\lambda \in \interval[open]{10^{-12}}{10^{-2}}$ (optimized in log-space). Bayesian optimization is initialized with 50 points in the hyperparameter space, chosen using a Latin hypercube design \cite{Tang2008}. Subsequently, BO chooses the next hyperparameters using the LCB criterion. The optimization was stopped when reaching a cross-validation error as good as that of the grid search, or when the Bayesian optimization had converged---with a criterion of $\epsilon = 10^{-3}$ on the L1 distance between subsequently evaluated hyperparameters from the GP model, normalized by their scale in the domain. 

For grid search, we discretize the parameter space. For  $N$ we take the sparse grid  $N \in \{50,100,200\}$ as the exact number of nodes is immaterial, e.g. we expect little difference between a reservoir of $55$ and one of $60$ nodes. 

For both $w_{in}$ and $w$ we take the same grid of $10$ equally spaced values starting with $0.01$ and ending with $0.95$. For regularization we take $5$ values for $\lambda \in \{10^{-12},\dots,10^{-2}\}$. This amounts to $1500$ parameter combinations.
Though we define the parameter grids manually, according to our a-priori expectation of how sensitive the individual parameters are, we note that this can be of tremendous benefit for the grid-search; if we were to uniformly discretize the parameter space in a grid of equally spaced grid points in each parameter dimension, grid search would perform very poorly.

Experiments were performed on the four datasets previously described.
Each experiment was repeated 30 times with different random seeds for both the grid search and the Bayesian optimization. All datasets were 
normalized to have mean zero and a standard deviation of $1$. 

The final score of each method was reported as the predictive performance on a held out test set and noted in Table \ref{table:results}. Test errors are reported as Normalized Mean Square Error (NMSE), which is defined the mean square error normalized by the series standard deviation:
\begin{equation}
\small
	\mathrm{NMSE} = \frac{\sum_{t=1}^{T} (\mathbf{y}(t) - \mathbf{\hat{y}}(t))^2}{\sum_{t=1}^{T} (\mathbf{y}(t) - \mathbf{\bar{y}}(t))^2} \ ,
\end{equation}
where, as in section \ref{background}, $\mathbf{y}(t)$ is the observed time series we want to predict at time step $t$, $\mathbf{\bar{y}}(t)$ is the empirical mean of the observed time series and $\mathbf{\hat{y}}(t)$ is the prediction of the SCR.
The NMSE facilitates an easier comparison of the results across datasets as it does not depend on the scaling  of the data.
\begin{table}
    \small
   	\centering
    \caption{Mean number of model evaluations needed for Bayesian Optimization to match the performance of a grid search. Sample standard deviations noted between parentheses. (30 repetitions)}
	\begin{tabular}{@{\hskip2pt}l@{\hskip4pt}c@{}c@{\hskip4pt}c@{\hskip4pt}c@{\hskip2pt}}
      \toprule
      {} & \multicolumn{2}{c}{Bayesian Optimization} & \multicolumn{2}{c}{Grid Search} \\ 
      \cmidrule(lr){2-3} 
      \cmidrule(lr){4-5}
      Dataset  & Evals & Test NMSE & Evals & Test NMSE \\ 
      \midrule
      \multirow{2}{*}{Mackey-Glass} & \textbf{89.7} 		& 0.0325 	& 1500 	& 0.0334 \\
                                    & (18.7) 			& (0.0044) 	& (0) 	& (0.0045) \\
      \multirow{2}{*}{NARMA}	 		& \textbf{143.4}   	& 0.0065	 	& 1500 	& 0.0065 \\ 
                                    & (56.6) 			& (0.0003) 	& (0) 	& (0.0001) \\
      \multirow{2}{*}{EMG}  			& \textbf{185.4}		& 0.3206 	& 1500 	& 0.3265 \\
                                    & (74.6) 			& (0.0229) 	& (0) 	& (0.0055) \\
 \multirow{2}{*}{RRL Light Curve}  	& \textbf{194.6}		& 0.0028 	& 1500 	& 0.0029 \\
                                    & (24.2) 			& (0.0003) 	& (0) 	& (5.29e-5) \\
      \bottomrule
	\end{tabular}
  \label{table:results}
\end{table}
\subsubsection{Results} We present the results in table \ref{table:results}. Bayesian optimization  needs only a fraction of the number of model evaluations to reach a test error that is comparable to that of the grid search on all four datasets. The speed up in terms of model evaluations (choosing hyperparameters, building and training a model and reporting validation performance) is between 7.5 to 16 times, which is a significant computational gain. Note that this speedup could have been even higher when compared to an uninformed, uniform grid search over the space, because such a strategy does not exploit heuristics that are known to deliver good scores.

\subsection{Clustering Families of Time Series}

A goal of this work is to avoid the computational effort
in learning one ESN per time series in the dataset as this involves tuning both the readout weights as well as the SCR hyperparameters.
In this section we demonstrate how the cluster formulation can help in reducing the number of ESN models required for obtaining good predictive performance on synthetic and real-life datasets of astronomical light curves.
We emphasize that the goal of the clustering here is not to necessarily cluster the time series according to some existing class membership, but rather to find ESNs that display good predictive performance
for multiple time series and hence avoid the need for learning an individual ESN per time series, which can be very costly.
Of course, if time series of the same class exhibit similar temporal behavior, as a byproduct, they may end up being assigned to the same ESN cluster.

\subsubsection{Setup}
In the first experiment we generate 40 time series with 750 points each, from four distinct sources:
\begin{itemize}
	\item a 10th-order NARMA system
	\item a 20th-order NARMA system
	\item a Mackey-Glass DDE with $\tau = 17$
	\item a Mackey-Glass DDE with $\tau = 30$
\end{itemize}
These series are all highly nonlinear. Since there is resemblance in the series' time lags as well as in their definitions, clustering these series by their source is an especially hard task.

The second experiment is a real-life application of time series clustering using cluster-based ESNs. We show that we can effectively cluster three distinct classes of light curves from astronomy, without prior knowledge of their labels. This experiment includes 30 series from three different astronomical light curves, 10 for each type:
\begin{itemize}
	\item RR Lyrae (RRLs)
	\item Eclipsing Binaries (EBs)
	\item Quasar
\end{itemize}
The RRLs and EBs are periodic stars \cite{RRL2015}, while Quasars exhibit stochastic behavior \cite{Andrae2013}. Every series is binned to have 500 points, and RRLs and EBs are folded to have 10 periods. Missing values are interpolated quadratically and the series are smoothed by applying a Savitzky-Golay filter \cite{Savitzky1964}. All data is centered and standardized before clustering.

\subsubsection{Results}
\begin{figure}
	\centering
	\includegraphics{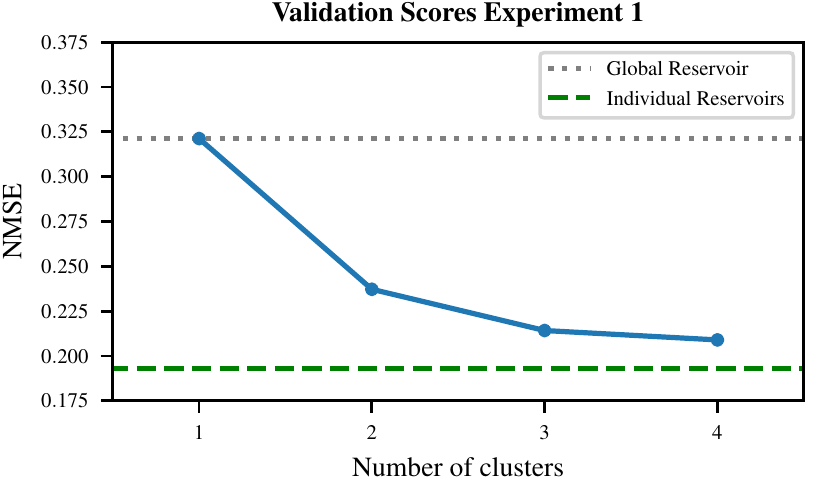}
	\caption{Validation error for our synthetic dataset of 40 series from four chaotic sources. At 4 clusters, predictive error converges to an error nearly equivalent of having one individual ESNs per time series (40 individual ESNs).}
	\label{fig:cluster}
\end{figure}
Fig. \ref{fig:cluster} shows how the mean predictive error changes given a number of clusters (models) for experiment 1. The dotted line denotes the average predictive performance of one global ESN with one readout and one set of hyperparameters (`one cluster'). As number of clusters increases, predictive error goes down rapidly, until adding more clusters does not significantly decrease predictive performance anymore. The predictive performance at 4 clusters is close to the predictive error of having 40 ESNs, or one for every individual time series. This shows that  the cluster-based ESNs represent the time series in their clusters well.

\begin{figure}
	\centering
	\includegraphics{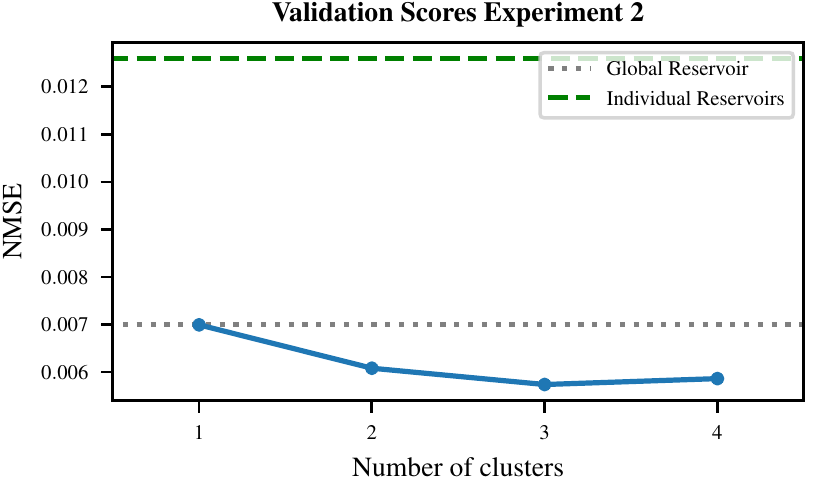}
	\caption{Validation error for our light curve dataset of 30 series from three stellar sources. Due to short sequences, modeling individual ESNs actually performs worse than modeling the series group-wise.}
	\label{fig:cluster_exp2}
\end{figure}
A similar plot is given for experiment 2 (Fig. \ref{fig:cluster_exp2}). The validation error reaches its minimum at 3 clusters and above. What is especially interesting is that due to the relatively short sequence length, individually optimized ESNs perform worse than one global ESN over all 30 series. We think this is due to overfitting, because, per sequence, effectively only 400 points are available for training and 100 points for validation. Combining series into clusters, therefore, groups and augments the data for each model, thereby making them less prone to overfitting.

For larger datasets the same approach can be followed. The number of clusters can be increased until a desired degree of predictive error is reached, or until some predetermined computation budget is exhausted. Alternatively, expert opinion may inform the number of clusters to be in accordance with the number of latent classes expected in the dataset.

\section{Conclusion} \label{conclusion}

We reduce the burden of ESN optimization by applying Bayesian optimization to ESNs hyperparameters, and have shown how to make effective use of it in a time series clustering context. Using these clusters we were able to reduce the amount of optimization needed even more, by precluding the need to model every individual series separately. In large time series dataset with redundancy, in which there often is only a limited number of data generating processes, only the clusters need to be modeled and optimized, leading to significant computational gains.

\bibliographystyle{IEEEtran}
\bibliography{IEEEabrv,library.bib}
\end{document}